\documentclass[11pt,a4paper]{article}
\usepackage[hyperref]{acl2020}
\usepackage{times}
\usepackage{latexsym}

\usepackage{graphicx}
\usepackage{extarrows}
\usepackage{amssymb}
\usepackage{multirow}
\usepackage{subfigure}

\usepackage{microtype}

\aclfinalcopy 

\title{CoKE: Contextualized Knowledge Graph Embedding}

\author{Quan Wang$^\ast$, Pingping Huang\thanks{\hspace{0.15cm}Equal contribution.} , Haifeng Wang, Songtai Dai, Wenbin Jiang  \\
{\bf Jing Liu, Yajuan Lyu, Yong Zhu, Hua Wu} \\
  Baidu Inc., Beijing, China \\
  {\tt \{wangquan05,huangpingping,wanghaifeng,daisongtai,jiangwenbin,}\\
  {\tt liujing46,lvyajuan,zhuyong,wu\_hua\}@baidu.com}}

\date{}

\begin{document}
\maketitle
\begin{abstract}
Knowledge graph embedding, which projects symbolic entities and relations into continuous vector spaces, is gaining increasing attention. Previous methods allow a single static embedding for each entity or relation, ignoring their intrinsic contextual nature, i.e., entities and relations may appear in different graph contexts, and accordingly, exhibit different properties. This work presents \textit{\underline{Co}ntextualized \underline{K}nowledge} \textit{Graph \underline{E}mbedding} (CoKE), a novel paradigm that takes into account such contextual nature, and learns dynamic, flexible, and fully contextualized entity and relation embeddings. Two types of graph contexts are studied: edges and paths, both formulated as sequences of entities and relations. CoKE takes a sequence as input and uses a Transformer encoder to obtain contextualized representations. These representations are hence naturally adaptive to the input, capturing contextual meanings of entities and relations therein. Evaluation on a wide variety of public benchmarks verifies the superiority of CoKE in link prediction and path query answering. It performs consistently better than, or at least equally well as current state-of-the-art in almost every case, in particular offering an absolute improvement of 21.0\% in H@10 on path query answering. Our code is available at {\small \url{https://github.com/PaddlePaddle/Research/tree/master/KG/CoKE}}.
\end{abstract}

\section{Introduction}
Recent years have seen rapid progress in knowledge graph (KG) construction and application. A KG is typically a multi-relational graph composed of entities as nodes and relations as different types of edges. Each edge is represented as a subject-relation-object triple $\left(s, r, o \right)$, indicating a specific relation between the two entities. Although such triples are effective in organizing knowledge, their symbolic nature makes them difficult to handle by most learning algorithms. KG embedding, which aims to project symbolic entities and relations into continuous vector spaces, has thus been proposed and quickly gained broad attention \cite{nickel2016:review,wang2017:review}. These embeddings preserve the inherent structures of KGs, and have shown to be beneficial in a variety of downstream tasks, e.g., relation extraction \cite{weston2013:RE,riedel2013:UniversalSchemas} and question answering \cite{bordes2014:QA-SubgraphEmbedding,yang2019:ktnet}.

\begin{figure}[!t]
	\centering
	\includegraphics[width=0.4 \textwidth]{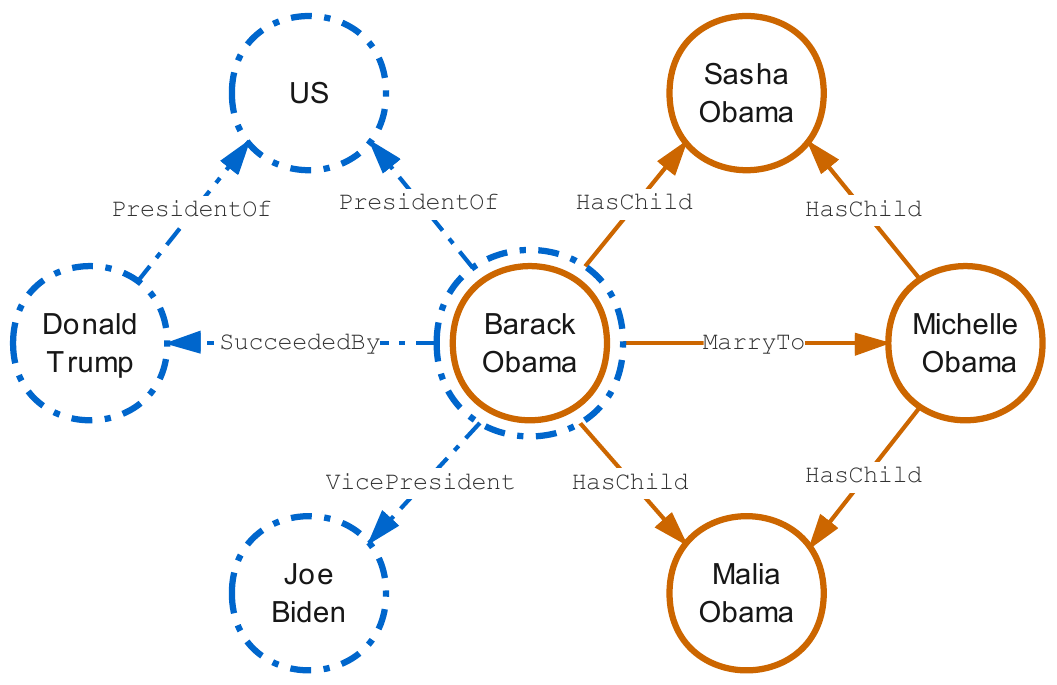}\\
	\caption{An example of \texttt{\small BarackObama}, where the left subgraph shows his political role (dashed blue) and the right one his family role (solid orange).}
	\label{fig:obama}
\end{figure}

Current approaches typically learn for each entity or relation a single static representation, to describe its global meaning in a given KG. However, entities and relations rarely appear in isolation. Instead, they form rich, varied graph contexts such as edges, paths, or even subgraphs. We argue that entities and relations, when involved in different graph contexts, might exhibit different meanings, just like words do when they appear in different textual contexts~\cite{peters2018:elmo}. Figure~\ref{fig:obama} provides an example of entity \texttt{BarackObama}. The left subgraph (dashed blue) shows his political role as a former president of US, while the right one (solid orange) shows his family role as a husband and a father, which possess quite different properties. Take relation \texttt{HasPart} as another example, which also presents contextualized meanings, e.g., composition-related as (\texttt{Table}, \texttt{HasPart}, \texttt{Leg}) and location-related as (\texttt{Atlantics}, \texttt{HasPart}, \texttt{NewYorkBay})~\cite{xiao2016:transg}. Learning entity and relation representations that could effectively capture their contextual meanings poses a new challenge to KG embedding.

Inspired by recent advances in contextualized word embedding~\cite{devlin2018:bert}, we propose \textit{\underline{Co}ntextualized \underline{K}nowledge Graph \underline{E}mbedding} (or CoKE for short), a novel KG embedding paradigm that is flexible, dynamic, and fully contextualized. Unlike previous methods that allow a single static representation for each entity or relation, CoKE models that representation as a function of input graph contexts. Two types of graph contexts are considered: edges and paths, both formalized as sequences of entities and relations. Given an input sequence, CoKE employs a stack of Transformer \cite{vaswani2017:transformer} blocks to encode the input and obtain contextualized representations for its components. The model is then trained by predicting a missing component in the sequence, based on these contextualized representations. In this way, CoKE learns KG embeddings dynamically adaptive to each input sequence, capturing contextual meanings of entities and relations therein.

We evaluate CoKE with two tasks: link prediction and path query answering~\cite{guu2015:pathquery}. Both can be formulated exactly in the same way as how CoKE is trained, i.e., to predict a missing entity from a given sequence (triple or path). CoKE performs extremely well in these tasks. It outperforms, or at least performs equally well as current state-of-the-art in almost every case. In particular, it offers an absolute improvement of up to 21.0\% in H@10 on path query answering, demonstrating its superior capability for multi-hop reasoning. Though using Transformer, CoKE is still parameter efficient, achieving better or comparable results with even fewer parameters. Visualization further demonstrates that CoKE can discern fine-grained contextual meanings of entities and relations.

We summarize our contributions as follows: (1) We propose the notion of contextualized KG embedding, which differs from previous paradigms by modeling contextual nature of entities and relations in KGs. (2) We devise a new approach CoKE to learn fully contextualized KG embeddings. We show that CoKE can be naturally applied to a variety of tasks like link prediction and path query answering. (3) Extensive experiments demonstrate the superiority of CoKE. It achieves new state-of-the-art results on a number of public benchmarks.

\section{Related Work}
KG embedding aims at learning distributed representations for entities and relations of a given KG. Recent years have witnessed increasing interest in this task, and various KG embedding techniques have been devised, e.g., translation-based models \cite{bordes2013:transe,wang2014:transh,lin2015:transr}, simple semantic matching models \cite{yang2015:distmult,nickel2016:hole,trouillon2016:complex}, and neural network models \cite{dettmers2018:conve,jiang2019:convr,nguyen2018:convkb}. We refer readers to~\cite{nickel2016:review,wang2017:review} for a thorough review. Most of these traditional models learn a static, global representation for each entity or relation, solely from individual subject-relation-object triples. 

Beyond triples, recent work tried to use more global graph structures like multi-hop paths \cite{lin2015:ptranse,das2017:chains} and $k$-degree neighborhoods \cite{feng2016:gake,schlichtkrull2017:rgcn} to learn better embeddings. Although such approaches take into account rich graph contexts, they are not ``contextualized'', still learning a static global representation for each entity/relation. 

The contextual nature of entities and relations has been noted previously, but from distinct views. Consider a classic translation-based model TransE \cite{bordes2013:transe}. To overcome its disadvantages in dealing with 1-to-N, N-to-1 and N-to-N relations, some researchers introduced relation-specific projections, by which an entity would get different projected representations when involved in different relations~\cite{wang2014:transh,lin2015:transr,ji2015:transd}. \citeauthor{xiao2016:transg}~\shortcite{xiao2016:transg} noted that relations can be polysemous, showing different meanings with different entity pairs. So they modeled relations as mixtures of Gaussians to deal with this polysemy issue. Although similar phenomena have been touched upon in previous work, there is little formal discussion about the contextual nature of KGs, and the solutions, of course, are not ``contextualized''.

This work is inspired by recent advances in learning contextualized word representations \cite{mccann2017:cove,peters2018:elmo,devlin2018:bert}, by drawing connections of graph edges/paths to natural language phrases/sentences. Such connections have been studied extensively in graph embedding~\cite{perozzi2014:deepwalk,grover2016:node2vec,ristoski2016:rdf2vec,cochez2017:rdfglove}. But most of these approaches obtain static embeddings via traditional word embedding techniques, and fail to capture the contextual nature of entities and relations.

\section{Our Approach}\label{sec:Approach}
\begin{figure*}[!t]
	\centering
	\includegraphics[width=0.82 \textwidth]{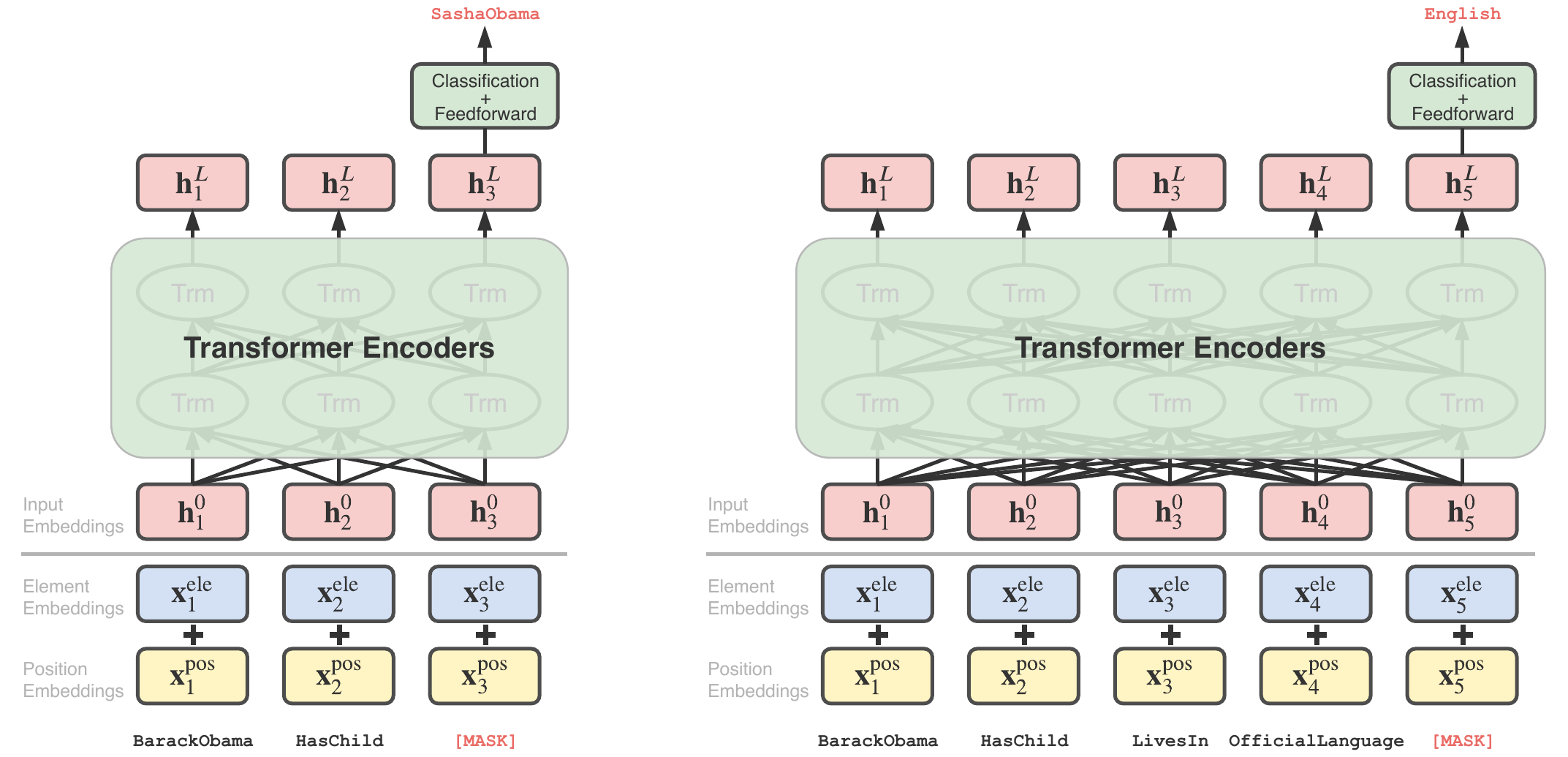}\\
	\caption{Overall framework of CoKE. An edge (left) or a path (right) is given as an input sequence, with an entity replaced by a special token \texttt{\small [MASK]}. The input is then fed into a stack of Transformer encoder blocks. The final hidden state corresponding to \texttt{\small [MASK]} is used to predict the target entity.}
	\label{fig:model}
\end{figure*}

Unlike previous methods that assign a single static representation to each entity/relation learned from the whole KG, CoKE models that representation as a function of each individual graph context, i.e., an edge or a path. Given a graph context as input, CoKE employs Transformer blocks to encode the input and obtain contextualized representations for entities and relations therein. The model is trained by predicting a missing entity in the input, based on these contextualized representations. Figure~\ref{fig:model} gives an overview of our approach. 

\subsection{Problem Formulation}\label{subsec:Formulation}
We are given a KG composed of subject-relation-object triples $\{(s,r,o)\}$. Each triple indicates a relation $r\in\mathcal{R}$ between two entities $s,o\in\mathcal{E}$, e.g., (\texttt{BarackObama}, \texttt{HasChild}, \texttt{SashaObama}). Here, $\mathcal{E}$ is the entity vocabulary and $\mathcal{R}$ the relation set. These entities and relations form rich, varied graph contexts. Two types of graph contexts are considered here: edges and paths, both formalized as sequences composed of entities and relations.
\begin{itemize}
\item An \textit{edge} $s \rightarrow r \rightarrow o$ is a sequence formed by a triple, e.g., \texttt{\small BarackObama} $\rightarrow$ \texttt{\small HasChild} $\rightarrow$ \texttt{\small SashaObama}. This is the basic unit of a KG, and also the simplest form of graph contexts. 
\item A \textit{path} $s \rightarrow r_1 \rightarrow \cdots \rightarrow r_k \rightarrow o$ is a sequence formed by a list of relations linking two entities, e.g., \texttt{\small BarackObama} $\rightarrow$ \texttt{\small HasChild} $\xlongrightarrow{(\texttt{\scriptsize Sasha})}$ \texttt{\small LivesIn} $\xlongrightarrow{(\texttt{\scriptsize US})}$ \texttt{\small OfficialLanguage} $\rightarrow$ \texttt{\small English}.\footnote{Entities in parentheses (\texttt{\small Sasha} and \texttt{\small US}) are not components of the path, just used to show how the path is generated.} The length of a path is defined as the number of relations therein. The example above is a path of length 3. Edges can be viewed as special paths of length 1.
\end{itemize}
Here we follow~\cite{guu2015:pathquery} and exclude intermediate entities from paths, by which the paths will get a close relationship with Horn clauses and first-order logic rules~\cite{lao2010:pra}. We leave the investigation of other path forms for future work. Given edges and paths that reveal rich graph structures, the aim of CoKE is to learn entity and relation representations dynamically adaptive to each input graph context.

\subsection{Model Architecture}\label{subsec:Model}
CoKE borrows ideas from recent techniques for learning contextualized word embeddings~\cite{devlin2018:bert}. Given a graph context, i.e., an edge or a path, we unify the input as a sequence $X=(x_1,x_2,\cdots,x_n)$, where the first and last elements are entities from $\mathcal{E}$, and the others in between are relations from $\mathcal{R}$. For each element $x_i$ in $X$, we construct its input representation as:
\begin{equation*}\label{eq:input}
\mathbf{h}_i^0 = \mathbf{x}_i^{\mathrm{ele}} + \mathbf{x}_i^{\mathrm{pos}},
\end{equation*}
where $\mathbf{x}_i^{\mathrm{ele}}$ is the element embedding and $\mathbf{x}_i^{\mathrm{pos}}$ the position embedding. The former is used to identify the current element, and the latter its position in the sequence. We allow an element embedding for each entity/relation in $\mathcal{E}\cup\mathcal{R}$, and a position embedding for each position within length $K$. 

After constructing all input representations, we feed them into a stack of $L$ successive Transformer encoders \cite{vaswani2017:transformer} to encode the sequence and obtain:
\begin{equation*}\label{eq:hidden}
\mathbf{h}_i^\ell = \mathrm{Transformer}(\mathbf{h}_i^{\ell-1}), \;\; \ell = 1, 2, \cdots, L,
\end{equation*}
where $\mathbf{h}_i^\ell$ is the hidden state of $x_i$ after the $\ell$-th layer. Unlike sequential left-to-right or right-to-left encoding strategies, Transformer uses a multi-head self-attention mechanism, which allows each element to attend to all elements in the sequence, and thus is more effective in context modeling. As the use of Transformer has become ubiquitous recently, we omit a detailed description and refer readers to \cite{vaswani2017:transformer} and supplementary material. The final hidden states $\{\mathbf{h}_i^L\}_{i=1}^{n}$ are taken as the desired representations for entities and relations within $X$. These representations are naturally contextualized, automatically adaptive to the input. 

\subsection{Model Training}\label{subsec:Training}
To train the model, we design an entity prediction task, i.e., to predict a missing entity from a given graph context. This task amounts to single-hop or multi-hop question answering on KGs.
\begin{itemize}
\item Each \textit{edge} $s \rightarrow r \rightarrow o$ is associated with two training instances: $? \rightarrow r \rightarrow o$ and $s \rightarrow r \rightarrow ?$. It is a single-hop question answering task, e.g., \texttt{\small BarackObama} $\rightarrow$ \texttt{\small HasChild} $\rightarrow ?$ is to answer ``\textit{Who is the child of Barack Obama?}''.
\item Each \textit{path} $s \rightarrow r_1 \rightarrow \cdots \rightarrow r_k \rightarrow o$ is also associated with two training instances, one to predict $s$ and the other to predict $o$. This is a multi-hop question answering task, e.g., \texttt{\small BarackObama} $\rightarrow$ \texttt{\small HasChild} $\rightarrow$ \texttt{\small LivesIn} $\rightarrow$ \texttt{\small OfficialLanguage} $\rightarrow ?$ is to answer ``\textit{What is the official language of the country where Barack Obama's child lives in?}''.
\end{itemize}
This entity prediction task resembles the masked language model (MLM) task studied in~\cite{devlin2018:bert}. But unlike MLM that randomly picks some input tokens to mask and predict, we restrict the masking and prediction solely to entities in a given edge/path, so as to create meaningful question answering instances. Moreover, many downstream tasks considered in the evaluation phase, e.g., link prediction and path query answering, can be formulated exactly in the same way as entity prediction (will be detailed in $\S$~\ref{sec:Exp}), which avoids training-test discrepancy. 

During training, for each edge or path unified as a sequence $X=(x_1,\cdots,x_n)$, we create two training instances, one by replacing $x_1$ with a special token \texttt{\small [MASK]} (to predict $s$), and the other by replacing $x_n$ with \texttt{\small [MASK]} (to predict $o$). Then, the masked sequence is fed into the Transformer encoding blocks. The final hidden state corresponding to \texttt{\small [MASK]}, i.e., $\mathbf{h}_1^L$ or $\mathbf{h}_n^L$, after a feedforward layer, is used to predict the target entity, via a standard softmax classification layer:
\begin{gather*}
\mathbf{z}_{1} = \mathrm{Feedforward}(\mathbf{h}_{1}^L), \; \mathbf{z}_{n} = \mathrm{Feedforward}(\mathbf{h}_{n}^L), \\
\mathbf{p}_{1} = \mathrm{softmax}(\mathbf{E}^{\mathrm{ele}} \mathbf{z}_{1}), \; \mathbf{p}_{n} = \mathrm{softmax}(\mathbf{E}^{\mathrm{ele}} \mathbf{z}_{n}).
\end{gather*}
Here, $\mathbf{z}_{1}$/$\mathbf{z}_{n}$ is the hidden state of $\mathbf{h}_1^L$/$\mathbf{h}_n^L$ after the feedforward layer, $\mathbf{E}^{\mathrm{ele}}\in\mathbb{R}^{V\times D}$ the classification weight shared with the input element embedding matrix, $D$ the hidden size, $V$ the entity vocabulary size, and $\mathbf{p}_{1}$/$\mathbf{p}_{n}$ the predicted distribution of $x_1$/$x_n$ ($s$/$o$) over all entities. Figure~\ref{fig:model} provides a visual illustration of this whole process.

We use cross-entropy between the one-hot label ($\mathbf{y}_1$/$\mathbf{y}_n$) and the prediction $\mathbf{p}_1$/$\mathbf{p}_n$ as training loss:
\begin{equation*}
\mathcal{L}(X)= -\sum\nolimits_t y_t \log p_t,
\end{equation*}
where $y_t$ and $p_t$ are the $t$-th components of $\mathbf{y}_1$/$\mathbf{y}_n$ and $\mathbf{p}_1$/$\mathbf{p}_n$ respectively. As one-hot labels restrict each entity prediction task to a single correct answer, we use a label smoothing strategy to lessen this restriction. We set $y_t=\epsilon$ for the target entity, and $y_t=\frac{1-\epsilon}{V-1}$ for each of the other entities. 

\section{Experiments}\label{sec:Exp}
We demonstrate the effectiveness of CoKE in link prediction and path query answering. We further visualize CoKE embeddings to show how they can discern contextual usage of entities and relations.

\subsection{Link Prediction}\label{subsec:LinkPrediction}
This task is to complete a triple $(s,r,o)$ with $s$ or $o$ missing, i.e., to predict $?\!\rightarrow
\! r \!\rightarrow\! o$ or $s\!\rightarrow\! r \!\rightarrow ?$ \cite{bordes2013:transe}. It is in the same form as our training task, i.e., entity prediction within edges.

\paragraph{Datasets}$\!\!$We conduct experiments on four widely used benchmarks. The statistics of the datasets are summarized in Table~\ref{tab:kbc-data}. FB15k and WN18 were introduced in~\cite{bordes2013:transe}, with the former sampled from Freebase and the latter from WordNet. FB15k-237 \cite{toutanova2015:nlfeat} and WN18RR \cite{dettmers2018:conve} are their modified versions, which exclude inverse relations and are harder to fit.

\begin{table}[!t]
\centering\small\setlength{\tabcolsep}{5pt}
\begin{tabular}{|l|r|r|r|r|}
        \hline
                         & FB15k & WN18 & FB15k-237 & WN18RR \\
        \hline\hline
        Entities     & 14,951       & 40,943   & 14,541   & 40,943    \\
        Relations  & 1,345         & 18          & 237        & 11           \\
        Train         & 483,142     & 141,442 & 272,115 & 86,835    \\
        Dev          & 50,000       & 5,000      & 17,535  & 3,034      \\
        Test          & 59,071       & 5,000      & 20,466  & 3,134      \\
        \hline
\end{tabular}
\caption{\label{tab:kbc-data} Number of entities, relations, and triples in each split of the four benchmarks.}
\end{table}

\paragraph{Training Details} 
In this task, we train our model with only \textit{triples} from the \textit{training} set. The maximum input sequence length is hence restricted to $K=3$. We use the following configuration for CoKE: the number of Transformer blocks $L=12$, number of self-attention heads $A=4$, hidden size $D=256$, and feed-forward size $2D=512$. We employ dropout on all layers, with the rate tuned in $\rho \in$ $\{.1, .2, .3, .4, .5\}$. The label smoothing rate is tuned in $\epsilon \in (0,1]$ with steps of $0.05$. We use the Adam optimizer \cite{kingma2014:adam} with a learning rate $\eta\!=\!5 \!\times\! 10^{-4}$. We also use learning rate warmup over the first 10\% training steps and linear decay of the learning rate. We train with batch size $B =512$ for at most 1000 epochs. The best hyperparameter setting on each dataset is determined by MRR (described later) on the dev set.

\paragraph{Evaluation Protocol}
During evaluation, given a test triple $(s,r,o)$, we replace $s$ with \texttt{\small [MASK]}, feed the sequence into CoKE, and obtain the predicted distribution of $s$ over all entities. We sort the distribution probabilities in descending order and get the rank of $s$. During ranking, we remove any $s'$ that $(s',r,o)$ already exists in the training, dev, or test set, i.e., a \textit{filtered} setting~\cite{bordes2013:transe}. This whole procedure is repeated while predicting $o$. We report the mean reciprocal rank (MRR) and the proportion of ranks no larger than $n$ (H@$n$).

\begin{table*}[t]
\centering\small\setlength{\tabcolsep}{6.5pt}
\begin{tabular}{|lccccccccc|}
\hline
\multirow{2}*{} & \multicolumn{4}{c}{FB15k} && \multicolumn{4}{c|}{WN18} \\
\cline{2-5} \cline{7-10}
& MRR & H@1 & H@3 & H@10 && MRR & H@1 & H@3 & H@10 \\
\hline\hline
\multicolumn{10}{|l|}{\it Methods that use triples alone} \\
SimplE~\cite{kazemi2018:simple}             & .727 & .660 & .773 & .838 && .942  & .939 & .944        & .947 \\
TorusE~\cite{ebisu2018:toruse}                 & .733 & .674 & .771 & .832 && .947 & .943 & .950        & .954 \\ 
ConvE~\cite{dettmers2018:conve}            & .745 & .670 & .801 & .873 && .942 & .935 & .947         & .955 \\
ConvR~\cite{jiang2019:convr}                   & .782 & .720 & .826 & .887 && .951 & .947 & {\bf .955} & .958 \\
RotatE~\cite{sun2019:rotate}                    & .797 & .746 & .830 & .884 && .949 & .944 & .952         & .959 \\ 
HypER~\cite{balavzevic2019:hyper}         & .790 & .734 & .829 & .885 && .951 & .947 & {\bf .955} & .958 \\
TuckER~\cite{balavzevic2019:tucker}       & .795 & .741 & .833 & .892 && {\bf .953} & {\bf .949} & {\bf .955} & .958 \\
\hline
\multicolumn{10}{|l|}{\it Methods that use graph contexts or rules} \\
R-GCN+~\cite{schlichtkrull2017:rgcn}       & .696 & .601 & .760 & .842 && .819 & .697 & .929 & {\bf .964} \\ 
KBLRN~\cite{garcia2017:kblrn}                 & .794 & .748 & --     & .875 && --      & --      & --     & -- \\
ComplEx-{\scriptsize NNE+AER}~\cite{ding2018:complex-nne-aer} & .803 & .761 & .831 & .874 && .943 & .940 & .945 & .948 \\
pLogicNet$^\ast$~\cite{qu2019:plogicnet} & .844 & .812 & .862 & .902 && .945 & .939 & .947 & .958 \\
\hline\hline
CoKE (with triples alone) & {\bf .855} & {\bf .826} & {\bf .872} & {\bf .906} && .952 & .947 & {\bf .955} & .960  \\
\hline
\end{tabular}
\caption{\label{tab:link-prediction} Link prediction results on FB15k and WN18. Baseline results are taken from original papers.}
\end{table*}

\begin{table*}[t]
\centering\small\setlength{\tabcolsep}{6.5pt}
\begin{tabular}{|lccccccccc|}
\hline
\multirow{2}*{} & \multicolumn{4}{c}{FB15k-237} && \multicolumn{4}{c|}{WN18RR} \\
\cline{2-5} \cline{7-10}
& MRR & H@1 & H@3 & H@10 && MRR & H@1 & H@3 & H@10 \\
\hline\hline
\multicolumn{10}{|l|}{\it Methods that use triples alone} \\
ConvE~\cite{dettmers2018:conve}            & .316 & .239 & .350 & .491 && .46~~  & .39~~ & .43~~  & .48~~ \\
ConvR~\cite{jiang2019:convr}                   & .350 & .261 & .385 & .528 && .475    & .443   & .489     & .537 \\
RotatE~\cite{sun2019:rotate}                    & .338 & .241 & .375 & .533 && .476    & .428   & .492     & {\bf .571} \\ 
HypER~\cite{balavzevic2019:hyper}         & .341 & .252 & .376 & .520 && .465    & .436   & .477     & .522 \\
TuckER~\cite{balavzevic2019:tucker}       & .358 & .266 & .394 & .544 && .470    & .443   & .482     & .526 \\
\hline
\multicolumn{10}{|l|}{\it Methods that use graph contexts or rules} \\
R-GCN+~\cite{schlichtkrull2017:rgcn}       & .249 & .151 & .264 & .417 && -- & -- & -- & -- \\ 
KBLRN~\cite{garcia2017:kblrn}                 & .309 & .219 & --      & .493 && -- & -- & -- & -- \\
pLogicNet$^\ast$~\cite{qu2019:plogicnet} & .332 & .237 & .367 & .524 && .441 & .398 & .446 & .537 \\
\hline\hline
CoKE (with triples alone) & {\bf .364} & {\bf .272} & {\bf .400} & {\bf .549} && {\bf .484} & {\bf .450} & {\bf .496} & .553  \\
\hline
\end{tabular}
\caption{\label{tab:link-prediction-hard} Link prediction results on FB15k-237 and WN18RR. Baseline results are taken from original papers.}
\end{table*}

\paragraph{Main Results}
Tables~\ref{tab:link-prediction} and~\ref{tab:link-prediction-hard} report link prediction results on the four datasets.  We select competitive baselines from the most recent publications with good results reported. Our baselines are categorized into two groups: methods that use triples alone and methods that further integrate rich graph contexts or logic rules (rules have a close relationship to multi-hop paths). CoKE falls into the first group as it uses only triples from the training set. 

We can see that CoKE outperforms all the competitive baselines on three out of the four datasets (FB15k, FB15k-237, and WN18RR) in almost all metrics, and obtains near the best results on WN18. CoKE is also the most stable among the methods. It performs consistently the best (or near the best) on all the datasets, while the baselines fail to do so (e.g., pLogicNet$^\ast$ which performs quite well on FB15k underperforms on FB15k-237/WN18RR; TuckER and RotatE which perform near the best on these two datasets obtain substantially worse results on FB15k). The results demonstrate the superiority of CoKE in single-hop reasoning.

\begin{table}[t]
\centering\small
\begin{tabular}{|l|r|r|}
\hline
& \multicolumn{1}{c|}{FB15k} & \multicolumn{1}{c|}{FB15k-237} \\
\hline\hline
RotatE    & 31.25M $|$ .797 $|$ .746 & 29.32M $|$ .338 $|$ .241 \\
TuckER   & 11.26M $|$ .795 $|$ .741 & 10.96M $|$ .358 $|$ .266 \\
CoKE      & 10.58M $|$ .855 $|$ .826 & 10.19M $|$ .364 $|$ .272 \\
\hline\hline
& \multicolumn{1}{c|}{WN18} & \multicolumn{1}{c|}{WN18RR} \\
\hline\hline
RotatE     & 40.95M $|$ .949 $|$ .944 & 40.95M $|$ .476 $|$ .428 \\
TuckER    &   9.39M $|$ .953 $|$ .949 &   9.39M $|$ .470 $|$ .443 \\
CoKE       & 16.92M $|$ .952 $|$ .947 & 16.92M $|$ .484 $|$ .450 \\
\hline
\end{tabular}
\caption{\label{tab:parameter-efficiency} Parameter efficiency on the four benchmarks. Each cell reports number of parameters, MRR, H@1.}
\end{table}

\paragraph{Parameter Efficiency}
We investigate parameter efficiency of CoKE. For comparison, we consider RotatE \cite{sun2019:rotate} and TuckER \cite{balavzevic2019:tucker}, which achieve previous state-of-the-art results with their optimal configurations explicitly stated. Table~\ref{tab:parameter-efficiency} presents the results on the four benchmarks. For each method, we report the number of parameters associated with the optimal configuration that leads to the performance shown in Tables \ref{tab:link-prediction} and~\ref{tab:link-prediction-hard}. 

Though a Transformer structure is used, CoKE is still parameter efficient, achieving substantially better results with even fewer parameters on FB15k/ FB15k-237, and better or comparable results with a relatively small number of parameters on WN18/ WN18RR. The reason is that besides Transformer, entity embeddings contribute a lot to the parameters due to a large vocabulary size. As entity embeddings are required by all the methods, their size becomes key to parameter efficiency. CoKE is able to work well with a small embedding size $D=256$ on all the datasets.

\subsection{Path Query Answering}\label{subsec:PathQuery}
This task is to answer path queries on KGs~\cite{guu2015:pathquery}. A path query $s\rightarrow r_1 \rightarrow \cdots \rightarrow r_k \rightarrow ?$ consists of an initial entity $s$ and a sequence of relations $r_1, \cdots,r_k$. The answer is an entity $o$ that can be reached from $s$ by traversing $r_1,\cdots,r_k$. For example, \texttt{\small BarackObama} $\rightarrow$ \texttt{\small HasChild} $\rightarrow$ \texttt{\small LivesIn} $\rightarrow ?$ is to ask ``\textit{Where does Obama's child live?}'', and the answer is \texttt{\small US}. It is also formulated the same as our training task: entity prediction within paths. 

\paragraph{Datasets}
We adopt the two datasets released by \citeauthor{guu2015:pathquery}~\shortcite{guu2015:pathquery}, created from WordNet and Freebase.\footnote{\url{https://www.codalab.org/worksheets/0xfcace41fdeec45f3bc6ddf31107b829f}} Triples of these two datasets are split into training and test sets, and paths have already been generated by random walks. Paths used for training are sampled from the graph composed of training triples alone, with the following procedure: (1) Uniformly sample a path length $k\in\{2,\!\cdots\!,5\}$ and a starting entity $s\in\mathcal{E}$. (2) Perform a random walk starting at $s$, continuing $k$ steps by traversing $r_1,$ $\cdots,r_k$, and reaching entity $o$. (3) Output a path $s$ $\rightarrow r_1$ $\rightarrow \cdots \rightarrow r_k \rightarrow o$. Paths of length 1 are not sampled, but constructed by directly adding training triples. Paths used for test are generated from the whole graph containing both training and test triples, with the same procedure. Test paths which also appear as training instances are removed. See \cite{guu2015:pathquery} for details of dataset construction. Table~\ref{tab:pathquery-data} summarizes statistics of the two datasets.\footnote{The statistics reported here are calculated directly from the released data, which are slightly different from the numbers reported in the paper~\cite{guu2015:pathquery}.}

\begin{table}[!t]
\centering\small\setlength{\tabcolsep}{7.5pt}
\begin{tabular}{|l|r|r|}
        \hline
         & WordNet & Freebase \\
        \hline\hline
        Entities         & 38,551      & 75,043 \\
        Relations      & 11             & 13 \\
        Train Triples & 110,361    & 316,232\\
        Dev Triples  & 2,602        & 5,908 \\
        Test Triples  & 10,462      & 23,733\\
        Train Paths  & 2,129,539 & 6,266,058 \\
        Dev Paths   & 11,277      & 27,163 \\
        Test Paths   & 46,577       & 109,557 \\
        \hline
\end{tabular}
\caption{\label{tab:pathquery-data} Number of entities, relations, triples and paths in each split of the two datasets.}
\end{table}

\paragraph{Training Details} 
In this task, we train our model with \textit{paths} from the \textit{training} set (triples are paths of length 1). The maximum input sequence length is hence restricted to $K=7$ (at most 5 relations between 2 entities). We use the same configuration for CoKE as in link prediction, with a dropout rate $\rho=0.1$ and a label smoothing rate $\epsilon=1$. We train with a learning rate of $5 \times 10^{-4}$ and a batch size of 2048 for at most 20 epochs. We select the best epoch according to dev MQ (detailed later). 

\paragraph{Evaluation Protocol}
We follow the same evaluation protocol of \cite{guu2015:pathquery}, to make our results directly comparable. Specifically, for each path query $s \rightarrow r_1 \rightarrow \!\cdots\! \rightarrow r_k \rightarrow ?$, we define: (1) candidate answers $\mathcal{C}$ that ``type match'', namely entities that participate in the final relation $r_k$ at least once, i.e., $\mathcal{C} \triangleq \{o | \exists s' \;\mathrm{s.t.} (s',r_k,o)\in\mathcal{G}\}$; (2) correct answers $\mathcal{P}$ that can be reached from $s$ by traversing $r_1,\cdots,r_k$, i.e., $\mathcal{P} \triangleq \{ o | \exists e_1, \cdots, e_{k-1}$ $\mathrm{s.t.} \; (s,r_1,e_1), \cdots, (e_{k-1}, r_k, o) \in \mathcal{G}\}$; (3) incorrect answers $\mathcal{N} \triangleq \mathcal{C} \setminus \mathcal{P}$. Here $\mathcal{G}$ is the whole graph composed of training and test triples. We replace the target entity $o$ with \texttt{\small [MASK]}, feed the sequence into CoKE, and get the predicted distribution of $o$ over all entities. We rank correct answer $o$ along with incorrect answers $\mathcal{N}$ according to the distribution probabilities in descending order, and compute the quantile as fraction of incorrect answers ranked after $o$. The quantile ranges from 0 to 1, with 1 being optimal. We report the mean quantile (MQ) aggregated over all test paths, and also the percentage of test cases with the correct answer ranked in the top 10 (H@10).\footnote{H@10 used here is slightly different from that used in link prediction. Here incorrect answers are restricted to be entities that ``type match''. But there is no such restriction in link prediction. See \cite{guu2015:pathquery} and their evaluation script for details: {\small \url{https://www.codalab.org/worksheets/0xfcace41fdeec45f3bc6ddf31107b829f}}.}

\begin{table}[t]
\centering\small\setlength{\tabcolsep}{5.5pt}
\begin{tabular}{|l|c|c|c|c|}
\hline
\multirow{2}*{} & \multicolumn{2}{|c|}{WordNet} & \multicolumn{2}{|c|}{Freebase} \\
\cline{2-3} \cline{4-5}
~ &  MQ  &  H@10~~ & MQ  &  H@10~~ \\
\hline\hline
Bilinear-{\scriptsize COMP}$^\dag$   & 0.894        & 0.543$^\ast$ & 0.835 & 0.421~~ \\
DistMult-{\scriptsize COMP}$^\dag$  & 0.904        & 0.311~~         & 0.848 & 0.386~~ \\
TransE-{\scriptsize COMP}$^\dag$    & 0.933        & 0.435~~        & 0.880 & 0.505~~ \\ 
Path-RNN$^\ddag$                            & {\bf 0.989} & --                   & --        & -- \\
ROP$^\sharp$                                    & --               & --                  & 0.907  & 0.567$^\ast$ \\
\hline\hline
CoKE {\scriptsize(PATHS $\leq 1$)} & 0.735 & 0.167~~ & 0.727 & 0.373~~ \\
CoKE {\scriptsize(PATHS $\leq 2$)} & 0.908 & 0.495~~ & 0.894 & 0.595~~ \\
CoKE {\scriptsize(PATHS $\leq 3$)} & 0.926 & 0.596~~ & 0.906 & 0.646~~ \\
CoKE {\scriptsize(PATHS $\leq 4$)} & 0.936 & 0.655~~ & 0.931 & 0.712~~ \\
CoKE {\scriptsize(PATHS $\leq 5$)} & 0.942 & {\bf 0.679}~~ & {\bf 0.950} & {\bf 0.777}~~ \\
\hline
\end{tabular}
\caption{\label{tab:path-query} Path query answering results on WordNet and Freebase, with $^\dag$ results taken from \cite{guu2015:pathquery}, $^\ddag$ from \cite{das2017:chains}, and $^\sharp$ from \cite{yin2018:rop}.}
\end{table}

\paragraph{Main Results}
Table~\ref{tab:path-query} reports the results of path query answering on the two datasets. As baselines, we choose compositional Bilinear, DistMult, and TransE devised by \citeauthor{guu2015:pathquery}~\shortcite{guu2015:pathquery}, which model paths by combining relations with additions and multiplications. We also compare with Path-RNN \cite{das2017:chains} and ROP \cite{yin2018:rop}, which combine relations with recurrent neural networks. We test our approach in five settings: CoKE {\scriptsize(PATHS $\leq k$)} for $k=1, \cdots, 5$, which means training with paths of length 1 to $k$. The $k=5$ setting enables a fair comparison with the baselines, while the $k<5$ settings actually use shorter paths for training.

As we can see, CoKE performs extremely well on this task. CoKE {\scriptsize(PATHS $\leq 5$)} outperforms all the baselines (except for MQ on WordNet), offering an absolute improvement in H@10 of up to 13.6\% on WordNet and that of up to 21.0\% on Freebase, compared against the current best-so-far (denoted by $\ast$). Notably, CoKE can achieve good results even if trained on relatively short paths. The $k\!=\!3$ setting on WordNet and $k\!=\!2$ on Freebase already outperform (in H@10) the baselines trained on longer paths of length up to 5. The performance grows significantly as the maximum path length $k$ increases. The results demonstrate the superior capability of CoKE to model composition patterns within paths so as to support multi-hop reasoning.

\begin{figure}[t]
\centering
\subfigure{
	\label{fig:WN-Path}
	\includegraphics[height=0.185 \textwidth]{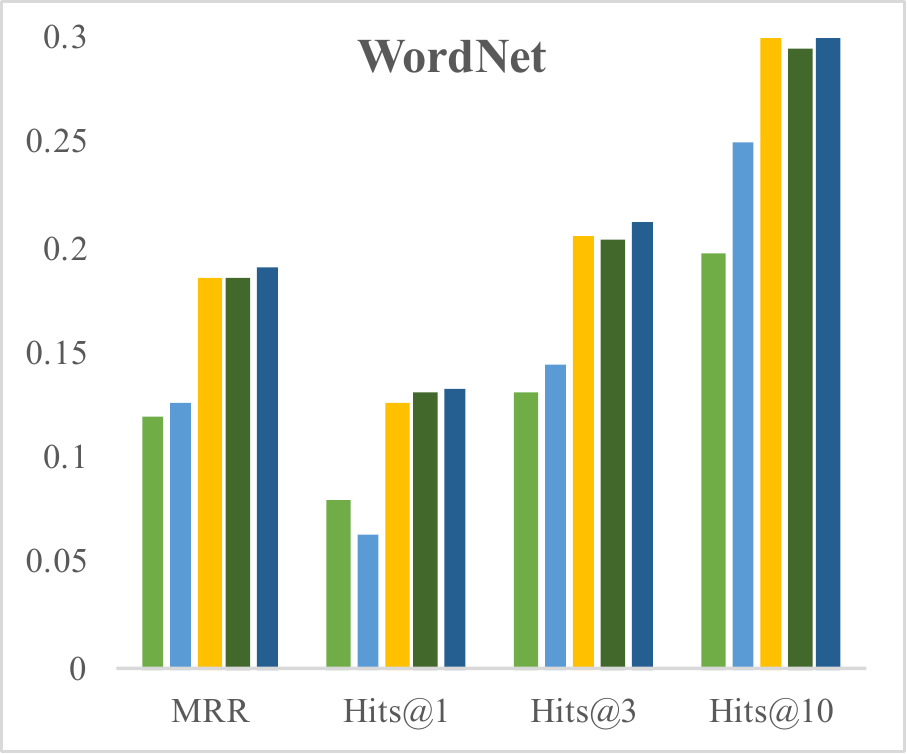}
}
\subfigure{
	\label{fig:FB-Path}
	\includegraphics[height=0.185 \textwidth]{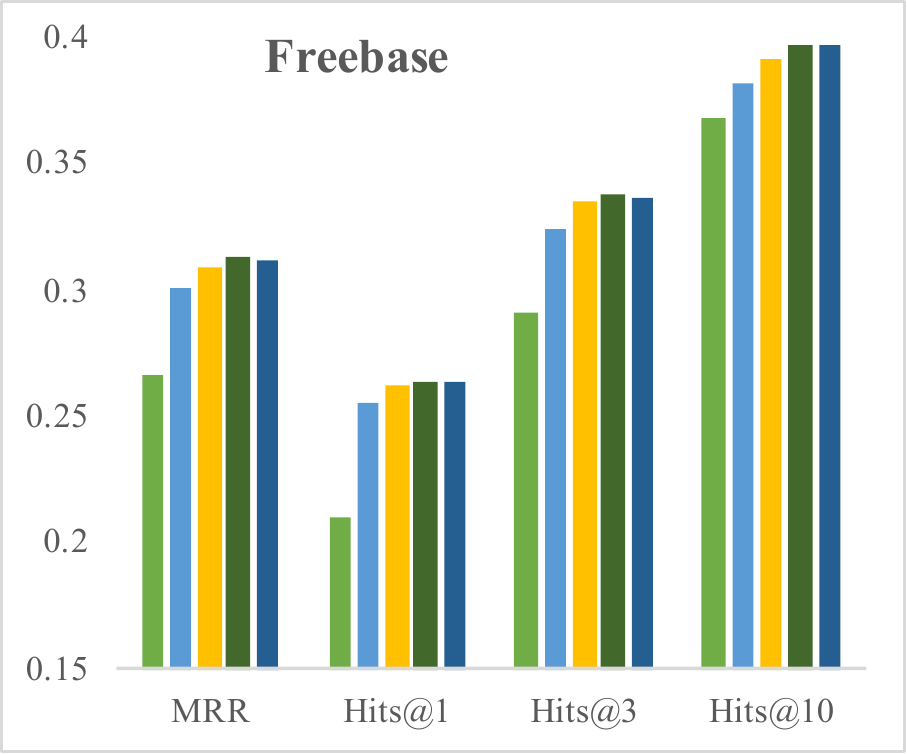}
}\\
\subfigure{
	\label{fig:label-Path}
	\includegraphics[width=0.45 \textwidth]{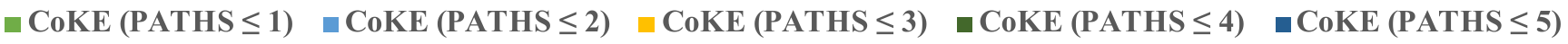}
}
\caption{Link prediction results on length-1 test paths from WordNet (left) and Freebase (right).}\label{fig:path-query-kbc}	
\end{figure} 

\begin{figure*}[t]
	\centering
	\includegraphics[width=1 \textwidth]{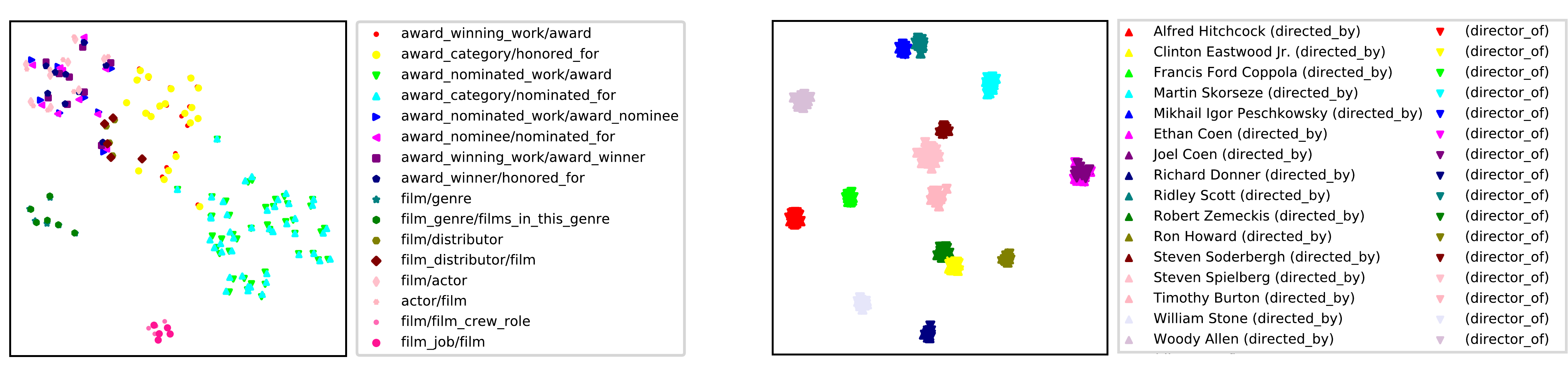}\\
	\caption{Contextualized representations of \texttt{\small TheKingsSpeech} (left) and \texttt{\small DirectorOf}/\texttt{\small DirectedBy} (right) learned by CoKE from FB15k. Each point is an entity/relation embedding within a triple. Different colors are used to distinguish different relations (left) or subjects/objects (right).}
	\label{fig:case_kingspeech}
\end{figure*} 

\paragraph{Further Analysis}
We further verify that training on multi-hop paths improves not only multi-hop reasoning but also single-hop reasoning. To do so, we consider a link prediction task on the two path query datasets. Specifically, we keep the training set unchanged (training paths of length 1 to 5), but consider only test paths of length 1. For each test triple $(s,r,o)$, we create two prediction cases: $?\!\rightarrow\!$ $r \!\rightarrow\! o$ and $s\!\rightarrow\! r \!\rightarrow ?$, and report aggregated MRR and H@$n$ for $n=1,3,10$ (see $\S$~\ref{subsec:LinkPrediction} for details).

We evaluate CoKE {\scriptsize(PATHS $\leq k$)} for $k=1,\cdots, 5$, which means training with paths of length up to $k$ but test only on paths of length 1. The results are presented in Figure~\ref{fig:path-query-kbc}. We can see that the $k\geq 2$ settings significantly outperform the $k=1$ setting in almost all metrics on both datasets (except for $k=2$ in H@1 on WordNet). And the performance generally grows as $k$ increases. The results verify that training on multi-hop paths further improves single-hop reasoning.

\subsection{Visual Illustrations}
This section provides visual illustrations of CoKE representations to show how they can distinguish contextual usage of entities and relations. 

We choose entity \texttt{\small TheKingsSpeech} from FB15k as an example, collecting all triples where it appears. We feed these triples into the optimal CoKE model learned during link prediction, and get final hidden states of this entity, i.e., its contextualized representations within different triples. We visualize these representations in a 2D plot via t-SNE \cite{van2008:t-sne}, and show the result in  Figure~\ref{fig:case_kingspeech} (left). Here, a different color is used for each relation, and relations appearing less than 5 times are discarded. We can see that representations of this entity vary across triples, falling into clusters according to the relations. Similar relations, e.g., \texttt{\small award\_winning\_work/award\_winner} and \texttt{\small award\_nominated\_work/award\_nominee}, tend to have overlapping clusters. This indicates the capability of CoKE to distinguish fine-grained contextual meanings of entities, i.e., how the meaning of an entity varies across relations. Moreover, we observe that the two representations, one obtained when this entity appears as a subject in $(s,r,o)$ and the other as an object in $(o,r^{-1},s)$, nearly coincide with each other in almost every case, where $r^{-1}$ is the inverse relation of $r$, e.g., \texttt{\small film/genre} and \texttt{\small genre/films\_in\_this\_genre}. This indicates that CoKE is pretty good at identifying relations and their inverse relations. 

Figure~\ref{fig:case_kingspeech} (right) further visualizes the contextualized representations of relation \texttt{\small DirectorOf} and its inverse relation \texttt{\small DirectedBy}, obtained in a similar way as the above case. Here, different colors are used to distinguish different directors. Directors appearing less than 10 times are discarded. Again, we observe that the two representations, one for \texttt{\small DirectorOf} in $(s,r,o)$ and the other for \texttt{\small Directed}  \texttt{\small By} in $(o,r^{-1},s)$, nearly coincide in almost every case. And these representations fall into clusters according to directors. The two overlapping clusters (rightmost ones) correspond to \texttt{\small JoelCoen} and \texttt{\small EthanCoen}, referred to as the Coen brothers who write, direct and produce films jointly. This indicates the capability of CoKE to distinguish fine-grained contextual meanings of relations, i.e., how the meaning of a relation varies across entities.

\section{Conclusion}\label{sec:Conclusion}
This paper introduces \textit{\underline{Co}ntextualized \underline{K}nowledge Graph \underline{E}mbedding} (CoKE), a novel paradigm that learns dynamic, flexible, and fully contextualized KG embeddings. Given an edge or a path formalized as a sequence of entities and relations, CoKE employs Transformer encoder to obtain contextualized representations for its components, which are naturally adaptive to the input, capturing contextual meanings of entities and relations therein. CoKE is conceptually simple yet empirically powerful, achieving new state of the art results in link prediction and path query answering on a number of widely used benchmarks. Visualization further demonstrates that CoKE representations can indeed discern fine-grained contextual meanings of entities and relations. 

As future work, we would like to (1) Generalize CoKE to other types of graph contexts beyond edges and paths, e.g., subgraphs of arbitrary forms. (2) Apply CoKE to more downstream tasks, not only those within a given KG, but also those scaling to broader domains. 

\bibliography{acl2020}
\bibliographystyle{acl_natbib}

\end{document}